\pdfoutput=1

\documentclass[11pt]{article}

\makeatletter
\def\@fnsymbol#1{\ensuremath{\ifcase#1\or \dagger\or *\or \ddagger\or
  \mathsection\or \mathparagraph\or \|\or **\or \dagger\dagger
  \or \ddagger\ddagger \else\@ctrerr\fi}}
\makeatother

\usepackage{graphicx}
\usepackage{booktabs}
\usepackage{enumitem}
\usepackage[]{acl}
\usepackage{amsmath}
\usepackage{amsfonts}
\usepackage{times}
\usepackage{latexsym}
\usepackage{subfigure}
\usepackage{tabularx}
\usepackage[T1]{fontenc}

\usepackage[utf8]{inputenc}

\usepackage{microtype}

\usepackage{algorithm}
\usepackage{algorithmic}

\usepackage{colortbl}

\usepackage{multirow}
\usepackage[normalem]{ulem}
\useunder{\uline}{\ul}{}

\usepackage{amssymb}
\usepackage{lipsum}
\usepackage{mathrsfs} 

\usepackage{longtable}

\usepackage{makecell}

\usepackage{colortbl}
\title{CNTLS: A Benchmark Dataset for Abstractive or  Extractive Chinese Timeline Summarization}

\author{\fontsize{11pt}{\baselineskip}\selectfont 
Qianren Mao\textsuperscript{$^{1,} \thanks{\quad These four authors contributed equally.}$},
Jiazheng Wang\textsuperscript{$^{2,} \footnotemark[1]$}, 
Zheng Wang\textsuperscript{$^{3,} \footnotemark[1]$}, 
Xi Li\textsuperscript{$^{2,} \footnotemark[1]$},
Bo Li\textsuperscript{$^{1,2}$}, 
Jianxin Li\textsuperscript{$^{1,2}$}
\\
 \fontsize{11pt}{\baselineskip}\selectfont \textsuperscript{$^{1}$}  Zhongguancun Laboratory, Beijing, P.R.China.\\
 \fontsize{11pt}{\baselineskip}\selectfont \textsuperscript{$^{2}$} School of Computer Science and Engineering, Beihang University, Beijing, P.R.China. \\
 \fontsize{11pt}{\baselineskip}\selectfont \textsuperscript{$^{3}$} Beijing-Dublin International College
 University College Dublin
 Belfield, Dublin, Ireland. \\
 \fontsize{11pt}{\baselineskip}\selectfont 
 \texttt{\{maoqr,libo,lijx\}@zgclab.edu.cn} \\
 \fontsize{11pt}{\baselineskip}\selectfont 
 \texttt{\{lixi,wangjzh\}@act.buaa.edu.cn}, \texttt{\{zheng.wang\}@ucdconnect.ie}
}

\begin{document}
\maketitle
\begin{abstract}
  Timeline summarization (TLS) involves creating summaries of long-running events using dated summaries from numerous news articles. However, limited data availability has significantly slowed down the development of timeline summarization. In this paper, we introduce the CNTLS dataset, a versatile resource for Chinese timeline summarization. CNTLS encompasses 77 real-life topics, each with 2524 documents and summarizes nearly 60\% days duration compression on average all topics. 
  We meticulously analyze the corpus using well-known metrics, focusing on the style of the summaries and the complexity of the summarization task. Specifically, we evaluate the performance of various extractive and generative summarization systems on the CNTLS corpus to provide benchmarks and support further research. To the best of our knowledge, CNTLS is the first Chinese timeline summarization dataset. The dataset and source code are released\footnote{Code and data available at: \emph{\url{https://github.com/OpenSUM/CNTLS}}.}. 
\end{abstract}


\section{Introduction}
With the rapid growth of web services, there is a continuous surge in the daily publication of news articles, covering a wide range of events from around the world. 
This sheer volume of news articles can overwhelm readers, making it challenging to navigate through this deluge of information.
To address this issue, it is necessary to develop techniques that help us tackle this huge amount of information. Timeline summarization (TLS) serves as a solution to the arduous manual summarization process, offering readers a faster and more comprehensive way to comprehend events from diverse viewpoints.
TLS is a technique designed to automatically extract sentences that depict the chronological progression of a particular topic from a large collection of web articles. This approach has garnered considerable attention in recent years~\cite{MartschatM18,GhalandariI20,QuatraCBMM21,DLiaoW021,YuJDSY20,MaoLWLHWW22,FaghihiAZRTJ22,YouLKFO22}.

Access to extensive and high-quality data is a fundamental prerequisite for significant advancements in the field of summarization. Particularly, timeline summarization presents even greater challenges due to the inclusion of informal and spoken expressions, frequent topic shifts, multiple participants, and extended context. This complexity makes the creation of large-scale, high-quality datasets for training neural summarization models a formidable task.
A practical TLS dataset should, at its core, mirror the statistical distribution of real major news events. Constructing a genuine timeline dataset entails the inclusion of a wide range of significant news events. Additionally, the generation of thousands of manual summaries or the development of new semi-automatic methods to obtain these summaries demands substantial human effort and innovation.
In summary, the availability of extensive and high-quality data plays a crucial role in advancing the field of summarization.

As in most NLP tasks, the majority of timeline summarization datasets, including TL17~\cite{TranAN13}, Crisis~\cite{TranHM15}, and Entities~\cite{GhalandariI20}, are predominantly available in English. The lack of comparable resources for other languages poses a challenge, limiting the potential impact of language technologies. Despite Chinese being the world's most widely spoken first language, used in 37 countries globally, it is often underrepresented or excluded in timeline summarization corpora.

In this paper, our primary goal is to create a high-quality, large-scale corpus suitable for training automatic timeline summarization in the Chinese language. To achieve this, we crawl Chinese timeline newspaper websites to construct annotation data in a straightforward manner.
These timeline summaries and articles, authored by professional writers and editors in Chinese newspapers, cover diverse topics such as news, sports, entertainment, finance, and more. They are provided as HTML metadata alongside articles, serving as page descriptions for news media services and search engines.
Crafted by human writers for general readers and explicitly designed for summarization, these Chinese newspaper summaries offer a detailed snapshot of how timeline summarization is applied in practice across various news topics across multiple documents.

The CNTLS corpus comprises 77 topics, each with its respective timeline summaries, providing extensive coverage surpassing most other English datasets. This abundance of data supports the evaluation of diverse event timeline methods, reducing dependency on specific datasets and enhancing result robustness. The CNTLS corpus includes articles and summaries spanning politics, economics, sports, culture, and various journalistic subjects. To our knowledge, CNTLS is the first timeline summarization dataset for the Chinese language.

To characterize the constructed corpus,  we employ four well-known metrics: extractive fragment coverage and density~\cite{GruskyNA18}, abstractivity\(_{p}\)~\cite{BommasaniC20}, and novel n-grams~\cite{KryscinskiPXS18}. For benchmarking, we assess automatic timeline summarization systems on the CNTLS corpus, including two major class of extractive systems (a clustering method of C{\footnotesize LUST}~\cite{GhalandariI20} and a data-ranking method of D{\footnotesize ATE}W{\footnotesize ISE}~\cite{GhalandariI20}).
The ORACLE system is used to compute upper bounds for extractive timeline summarization performance. Importanly, given the absence of generative methods in timeline summarization tasks and the current emergence of `large models', we also validate  the latest efficient generative pre-trained large language generative models (ChatGLM~\cite{DuQLDQY022}, Alpaca~\cite{Taori2023a}) capable of handling long-text inputs during summarization.

We evaluate the summarizers using four proprietary summarization metrics~\cite{MarkertM17} widely recognized for assessing timeline summary performance. These metrics cover concatenation-based Rouge F1 (Concat R1 or R2), date-agreement Rouge F1 (Agree R1 or R2), alignment-based Rouge F1 (align R1 or R2), and Date F1 score for data selection evaluation.

In summary, our contributions in this work include:
(i) Collecting a large and high-quality Chinese timeline summarization dataset from real-life news articles, namely CNTLS.
(ii) Conducting an analysis of the corpus using well-known metrics in the timeline summarization field, focusing on the character of the summaries and the difficulty of the timeline summarization task.
(iii) Evaluating the performance of extractive summarization systems on the CNTLS corpus for benchmarking purposes. Additionally, we pioneer the use of advanced large generative language models (fine-tuned from the original Meta's LLaMA or Stanford's Alpaca) for creating long timeline summaries.


\begin{table*}[htb]
  \footnotesize
  \centering
  \caption{Dataset Statistics for the English datasets and our Chinese CNTLS dataset.}
  \label{FFCC_RANK_results}
  \renewcommand\arraystretch{1.2}
  \setlength{\tabcolsep}{0.2mm}{
      \begin{tabular}{Xl|ccccccccccccccc}
      \toprule
  \textbf{Dataset}   & \texttt{{\tiny Topics/}}  & \texttt{{\tiny TLs/}}  & \texttt{{\tiny A.L/}} & \texttt{{\tiny A.DocN/}}  & \texttt{{\tiny A.Sent/}}  & \texttt{{\tiny A.Token/}} & \texttt{{\tiny A.DocL/}}  & \texttt{{\tiny A.DateT/}} & \texttt{{\tiny A.SumT/}} & \texttt{{\tiny A.K/}} & \texttt{{\tiny A.Duration/}} & \texttt{{\tiny A.DurComp/}}  & \texttt{{\tiny A.DateComp/}} & \texttt{{\tiny A.DateCov/}} \\ 
  \midrule
  \textbf{TL17}           & 9       & 19      & 36    & 467       & 20,409 & 945 
  & 441,346  & 990 & 4,561 &2.9  &212 &41.35  &0.45 &0.81  \\ 
  \textbf{Crisis}         & 4       & 22     & 29     & 4,393     & 82,761   &831   & 3,650,095 & 822 & 5,067 &1.3  &343 &19.50  &0.11 &0.9   \\ 
  \textbf{Entities}       & 47      & 47     & 23     & 959       & 31,545  &783 &840,655 & 793 &568  &1.2 &4,437 &0.48  & 0.06  &0.51 \\
  \textbf{CNTLS}          & 77      & 77   & 6   & 564    & 28,725 &590  &33,2361   & 1723  &698   &1.4   &55 &59.96  &0.51 &0.88 \\
  \bottomrule
  \end{tabular}%
  }
\end{table*}

\begin{table}[htb]
  \footnotesize
  \centering
  \caption{The Meaning of Abbreviated Symbols.}
  \label{FFCC_RANK_results}
  \renewcommand\arraystretch{1.2}
  \setlength{\tabcolsep}{5mm}{
      \begin{tabular}{@{}p{1.2cm}|p{4.7cm}}
      \toprule
  \textbf{Abbr}   & \textbf{Implication} \\ 
  \midrule
  \texttt{{\scriptsize Topics/}}          &  \textit{the number of topics of the dataset}.  \\ 
 \texttt{{\scriptsize TLs/}}        &  \textit{the number of ground truth timelines of \texttt{{\scriptsize Topics}}}.     \\ 
 \texttt{{\scriptsize A.L/}}     & \textit{the average number of daily sentences of \texttt{{\scriptsize Topics}}}.  \\
 \texttt{{\scriptsize A.DocN/}}         &  \textit{the average number of files for the single one of \texttt{\scriptsize Topics}}.   \\
   \texttt{{\scriptsize A.Sent/}}           & \textit{the average number of sentences in source articles for the single one of \texttt{\scriptsize Topics}}.   \\ 
  \texttt{{\scriptsize A.Token/}}         & \textit{the average number of word tokens for each source articles}      \\ 
 \texttt{{\scriptsize A.DocL/}}       & \textit{the average number of word tokens of all dates mentioned sentences for for the single one of \texttt{\scriptsize Topics}}.  \\
 \texttt{{\scriptsize A.DateT/}}          &  \textit{the average number of word tokens of all source articles for each date}   \\
 \texttt{{\scriptsize A.SumT/}}         & \textit{the average number of word tokens for the single one of \texttt{\scriptsize Topics}}.   \\ 
 \texttt{{\scriptsize A.K/}}         & \textit{the average number of sentences for the single one of \texttt{\scriptsize TLs}}.      \\ 
 \texttt{{\scriptsize A.Duration/}}      & \textit{the average number of days experienced from the beginning to the end of \texttt{\scriptsize TLs}}.  \\
 \texttt{{\scriptsize A.DurComp/}}         & \textit{the compression ratio w.r.t.\ timeline length is divided by duration}.   \\
   \texttt{{\scriptsize A.DateComp/}}         & \textit{the compression ratio w.r.t.\ dates is divided by the total number of dates mentioned in articles}.    \\ 
  \texttt{{\scriptsize A.DateCov/}}         &  \textit{the average coverage of dates in the ground truth timeline by the news in  articles collection}.     \\ 
  \bottomrule
  \end{tabular}%
  }
\end{table}

\section{Related Work}

\subsection{Timeline Summarization}
Since the inception of timeline summarization~\cite{SwanA00,AllanGK01}, this field has garnered considerable attention over the years~\cite{AlonsoGB09,YanKHWLZ11,ZhaoGYHL13,LiL13,TranAH15,WangMRS16,PasqualiCRSJJ21}.

To put it succinctly, the evolution of representative methods involves transitioning to either event clustering~\cite{AlonsoGB09,TranNKGA15,PasqualiMCJJ19,ZhaoLGJDZZH20,DuanJY20,GhalandariI20} or sentence ranking~\cite{DRadevJST04,LinB11,NguyenTM14a,Ghalandari17,GhalandariI20,MaoLWLHWW22} to select the optimal sentence.
~\citet{ChieuL04} first construct timelines by directly selecting the top-ranked sentences based on similarities within sentences and ~\citet{LiL13} first select dates then extract sentences corresponding to the dates. ~\citet{NguyenTM14a} propose a pipeline for generating timelines, involving date selection, sentence clustering, and sentence ranking. 
More recently, ~\citet{MartschatM18} adapt a submodular function model for the TLS task, originally used for multi-document summarization (MDS). Furthermore, ~\citet{GhalandariI20}  examine various TLS strategies and categorize TLS frameworks into three types: direct summarization approaches, date-wise approaches, and event detection approaches.

\subsection{Existing Datasets}
There are several frequently used timeline summarization datasets: TL17~\cite{TranAN13}, Crisis~\cite{TranHM15}, and Entities~\cite{GhalandariI20}. These datasets contain human-written timelines on specific topics, with source news articles retrieved from the web at a given point in time. Each dataset comprises journalist-generated timelines from major news media such as CNN, BBC, and Reuters, along with a corresponding corpus of articles per topic (e.g., H1N1 flu, Enron bankruptcy, and Egypt war).

Specifically, the number of topics and their time spans varies. TL17 contains 19 timelines from 9 topics, while Crisis involves 22 timelines from 4 topics. An overview of the existing English datasets is shown in Table~\ref{FFCC_RANK_results}.

\section{Building the CNTLS Corpus}

\subsection{Timeline Summary Scraping}
The CNTLS dataset is compiled from web media metadata through a web-scale crawl covering over 77 topics from various online publishers. We use the HTML crawl tool \texttt{BeautifulSoup}~\footnote{\emph{\url{https://www.crummy.com/software/BeautifulSoup/}}} to extract HTML body content from specific timeline newspaper websites (houxu.app/, dsj365.cn/, etc.). These websites specialize in organizing news topics, titles, and timeline summaries, providing access to explicit metadata of timeline summaries. The Chinese newspaper summaries found on these sites are authored by human writers, intended for general readership, and explicitly crafted for summarization purposes. We identify and utilize topics and their corresponding timeline summaries from the HTML metadata.

Receiving topics along with their corresponding timeline summary texts directly serves as annotation samples for creating timeline summaries. Each topic is associated with a timeline sequence displaying headlines with publication dates, and each headline has a corresponding summary.

\begin{table*}[htb]
  \centering
  \footnotesize
  \caption{Average values of the metrics in the Datasets.}
  \label{Average_metric_results}
  \renewcommand\arraystretch{1}
  \setlength{\tabcolsep}{0.72mm}{
      \begin{tabular}{Xl|cccccccccccccc}
      \toprule
  \textbf{Dataset}   &  \texttt{{\tiny Coverage/}}  & \texttt{{\tiny Density/}}  & \texttt{{\tiny Compression/}} & \texttt{{\tiny Abstractivity (p=2)/}}  & \texttt{{\tiny Novel 2-grams/}}  & \texttt{{\tiny Novel 3-grams/}} & \texttt{{\tiny Novel 4-grams/}} \\ 
  \midrule
  \textbf{TL17}           & 61.76       & 0.10    & 29.54    & 96.31      & 28.41 & 30.30  
  & 30.61  \\ 
  \textbf{Crisis}         & 41.57       & 0.08     & 16.00    & 97.26   &19.71   & 20.08 & 20.15  \\ 
  \textbf{Entities}       & 87.00      & 0.72     & 17.90    & 99.97       & 59.48  & 74.23 &90.84\\
  \textbf{CNTLS}          & 30.96      & 2.56   & 41.49   & 94.88    & 77.63 &82.60 &83.53\\
  \bottomrule
  \end{tabular}%
  }
\end{table*}

\subsection{Source Article Extraction}
To retrieve the original documents corresponding to the timeline summaries organized by dates, we follow the construction process outlined in the English timeline corpus~\cite{TranAN13,YanKHWLZ11}. We manually define a set of keywords for each topic. Initially, we use the \texttt{HanNLP} tool~\footnote{\emph{\url{https://github.com/hankcs/HanLP}}} to extract keywords from news headlines in each topic after removing stop words. Subsequently, we use the obtained start and end times as search criteria, along with the keywords from the headlines, to search for relevant documents in the RING~\cite{PengLSYRYH21} news database, stored in HBase and indexed using Elastic Search. This approach aims to retrieve multiple news articles for each known publication time, forming a set of news documents corresponding to each timeline summary. These news documents serve as input for the summarization system, producing a time-ordered list along with its corresponding summary.

For each document, we utilize \texttt{PyLTP}~\cite{abs-2009-11616}\footnote{\emph{\url{https://github.com/HIT-SCIR/pyltp}}} for Chinese text segmentation and we identify temporal expressions with \texttt{RecognizeTextDate}\footnote{\emph{\url{https://github.com/microsoft/Recognizers-Text}}}. If a document contains a mappable time expression, we assign the date of the source document accordingly (using the first expression if multiple, otherwise setting it to the publication date of the article).

\section{Analysis of CNTLS Dataset}
All ataset statistics are shown in Table~\ref{FFCC_RANK_results}. TL17 has longer (\texttt{{\small A.L}}=36) timelines than the other datasets, Crisis has more long sequence of word tokens for each timeline (\texttt{{\small A.DocL}}=3,650,095), and Entities has more extended periods (\texttt{{\small A.Duration}}=4437).

In contrast to other English datasets, the CNTLS dataset distinguishes itself with a significantly larger number of topics (\texttt{{\scriptsize Topics/}}). The average number of word tokens in input articles for each date (\texttt{{\scriptsize A.DateT/}}) is also notably larger than the other three English datasets.  Additionally, despite CNTLS having a relatively short timeline length (\texttt{{\scriptsize A.L/}}), its significant compression ratio of time duration (\texttt{{\scriptsize A.DurComp/}}) aligns closely with real-world scenarios.  Besides, this distinction sets it apart by offering a wealth of events and timelines, providing ample data for evaluating the generalizability of various timeline strategies and reducing the potential for results to be influenced by specific data. 

Besides, CNTLS exhibits the highest compression value (41.99), as shown in Table~\ref{Average_metric_results}, implying that the summarization system must compress a greater amount of original information to generate the summary during the timeline summarization process. Moreover, the number of novel tokens in the CNTLS dataset is also among the highest (novel 2-grams: 77.63 and novel 3-grams: 82.60) across several English datasets.

\subsection{Characterizing Summarization Metrics}
We analyze the abstractivity of the timeline summarization corpus using established metrics from prior works~\cite{GruskyNA18,SorianoAHG22} in dataset construction. These metrics gauge abstractivity by measuring the extent of text overlap between the summary and the article. Specifically, we employ the following metrics: Extractive Fragment Coverage and Density, Abstractivity\(_{p}\) and novel n-grams.
\textbf{Extractive Fragment Coverage}~\cite{GruskyNA18}: This metric quantifies the extent to which a summary is derived from the text, indicating the percentage of summary words belonging to extractive fragments of the article.  \textbf{Extractive Fragment Density}~\cite{GruskyNA18}: In contrast to coverage, density considers the length of extractive fragments. While high coverage may result from numerous individual words in the summary, low density suggests short extractive fragments. \textbf{Compression Ratio}~\cite{GruskyNA18}: This ratio measures the length ratio between the article and the summary, with higher compression presenting challenges in capturing critical aspects more precisely.
\textbf{Abstractivity}\(_{p}\)~\cite{BommasaniC20}: This metric quantifies abstractivity by assessing overlap between the summary and the original text. Higher values indicate reduced overlap, and the parameter \(p\) assigns weight to the length of each extractive fragment.
\textbf{Novel n-grams}~\cite{KryscinskiPXS18}: This metric quantifies n-grams introduced in the summary but absent in the original text. We explore novel n-grams without considering the generated timeline summarization, illustrating the intrinsic novel properties of datasets. The metric's value is expressed as a percentage of the total number of n-grams in the summary.

\begin{figure*}
  \centering
  \begin{minipage}[t]{0.24\linewidth}
    \centering
    \subfigure[TL17]{
      \includegraphics[width=1.5in]{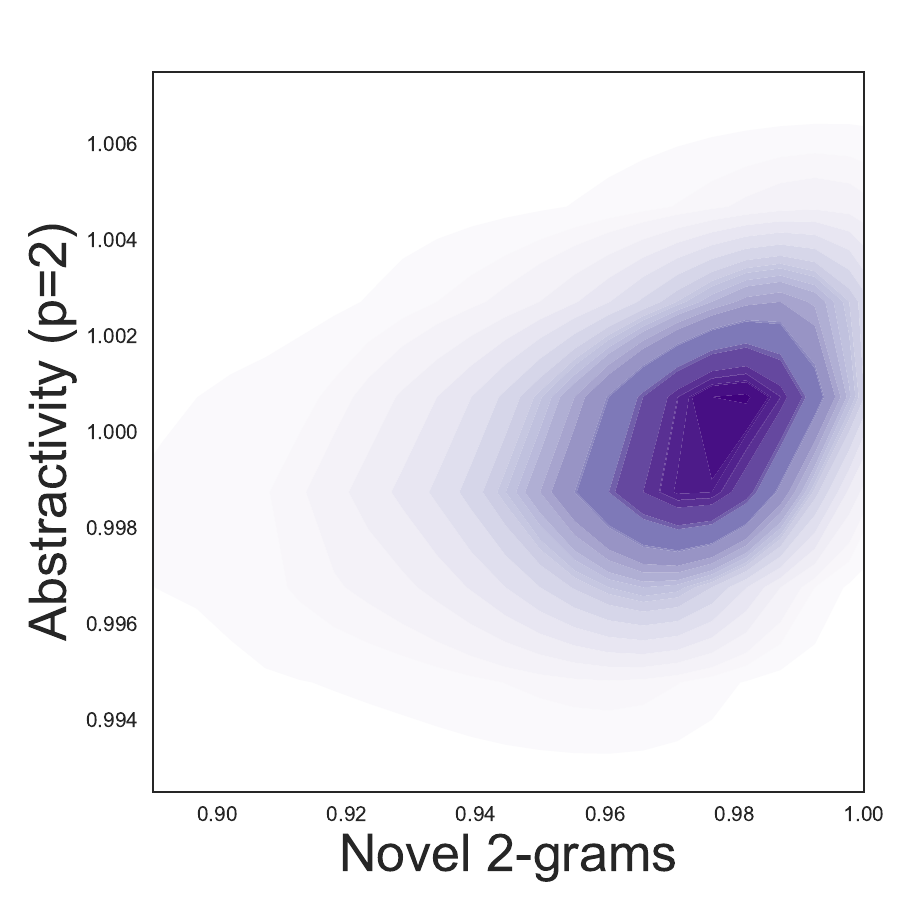}}
        \label{fig_side_GS}
    \end{minipage}%
      \begin{minipage}[t]{0.24\linewidth}
  \centering
  \subfigure[Crisis]{
    \includegraphics[width=1.5in]{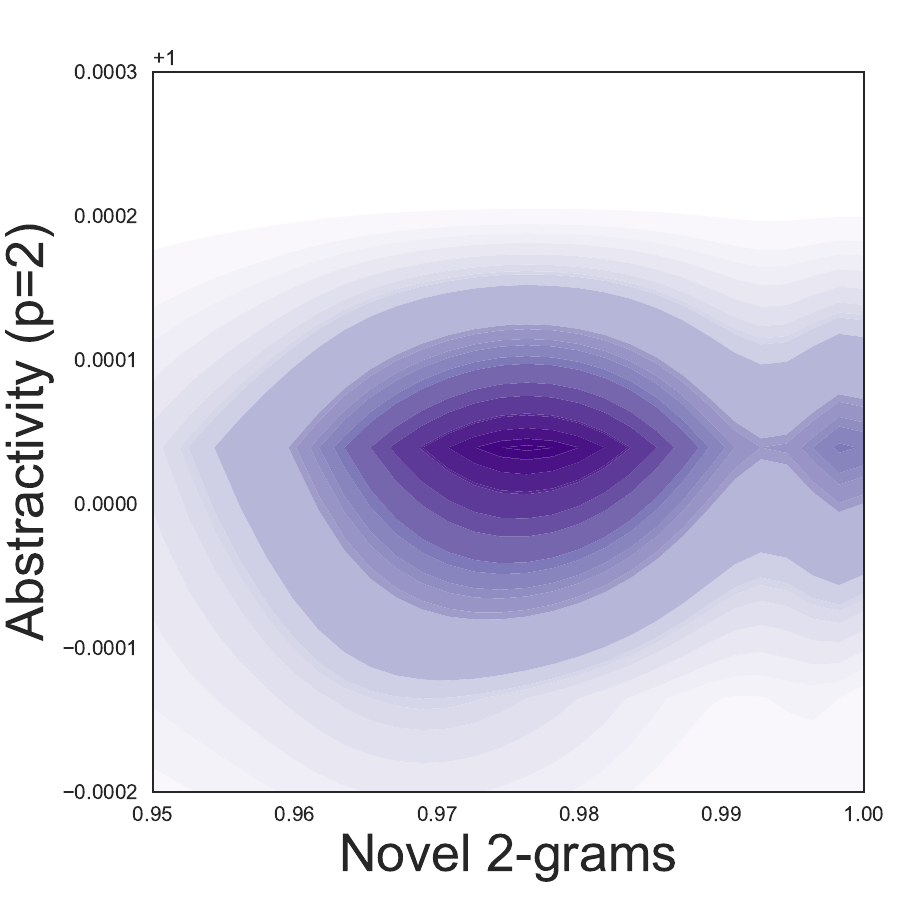}}
      \label{fig_side_GS}
  \end{minipage}%
     \begin{minipage}[t]{0.24\linewidth}
  \centering
      \subfigure[Entities]{
  \includegraphics[width=1.5in]{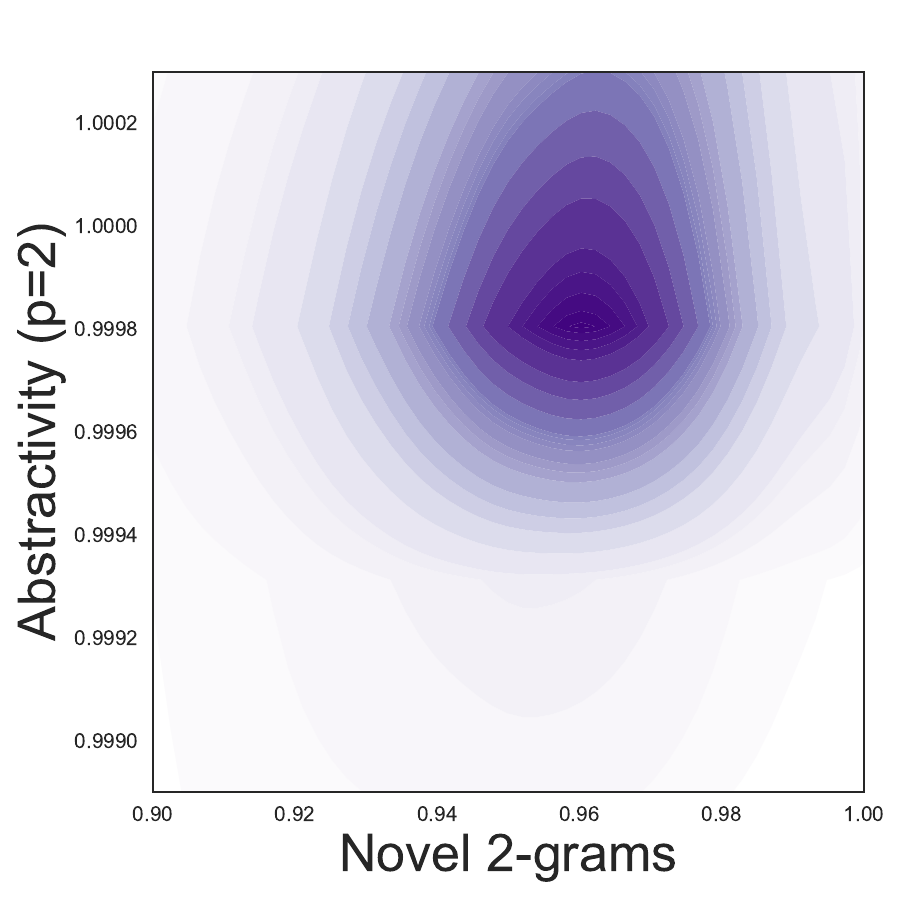}}
      \label{EGAT}
  \end{minipage}
    \begin{minipage}[t]{0.24\linewidth}
    \centering
        \subfigure[CNTLS]{
    \includegraphics[width=1.5in]{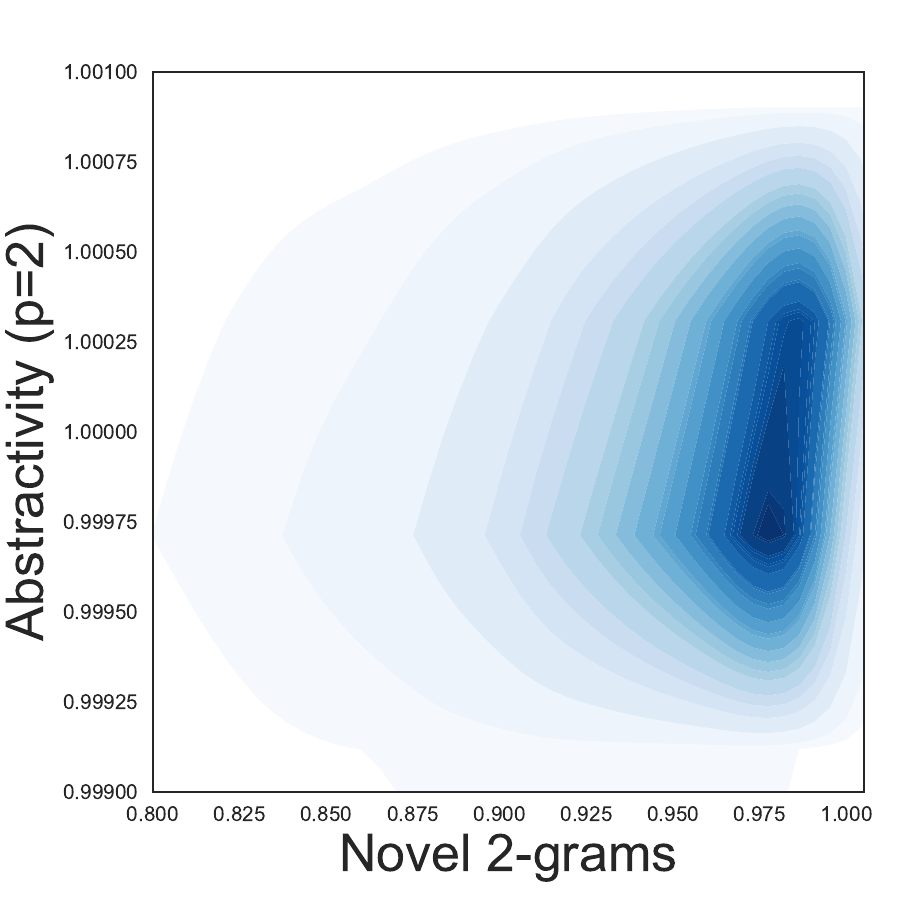}}
        \label{CNTLS1}
    \end{minipage}

    \vspace{-0.1in}
      \begin{minipage}[t]{0.24\linewidth}
        \centering
        \subfigure[TL17]{
          \includegraphics[width=1.5in]{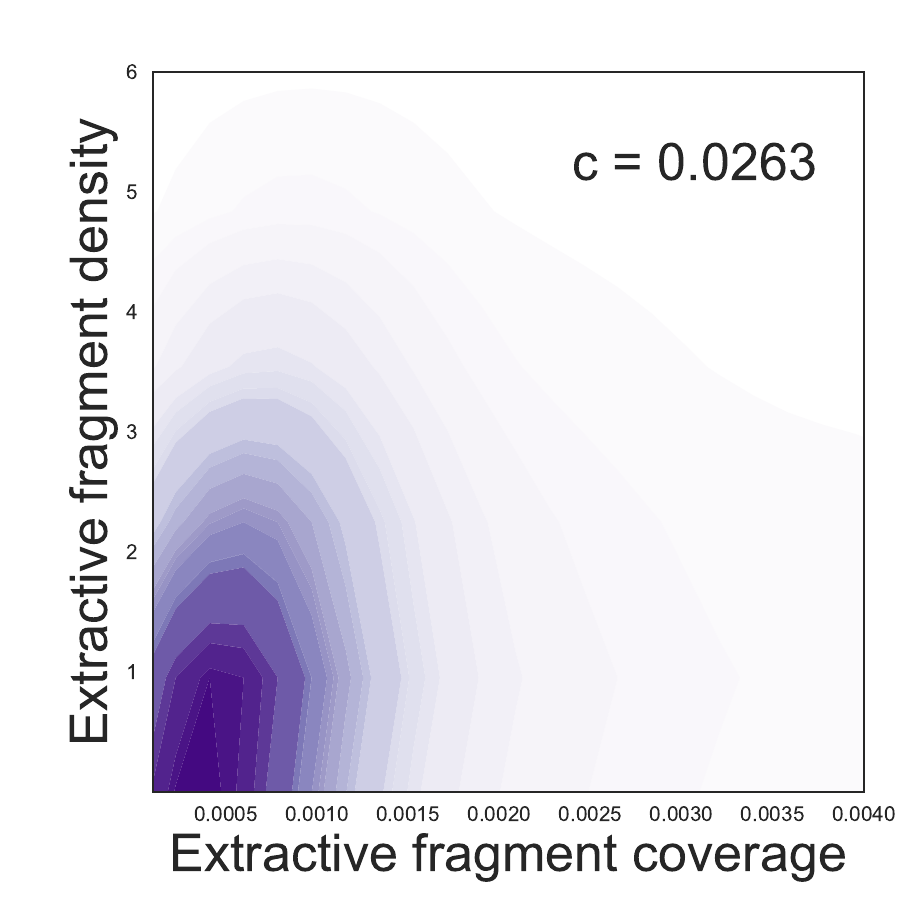}}
            \label{fig_side_GS}
        \end{minipage}%
    \begin{minipage}[t]{0.24\linewidth}
      \centering
      \subfigure[Crisis]{
        \includegraphics[width=1.5in]{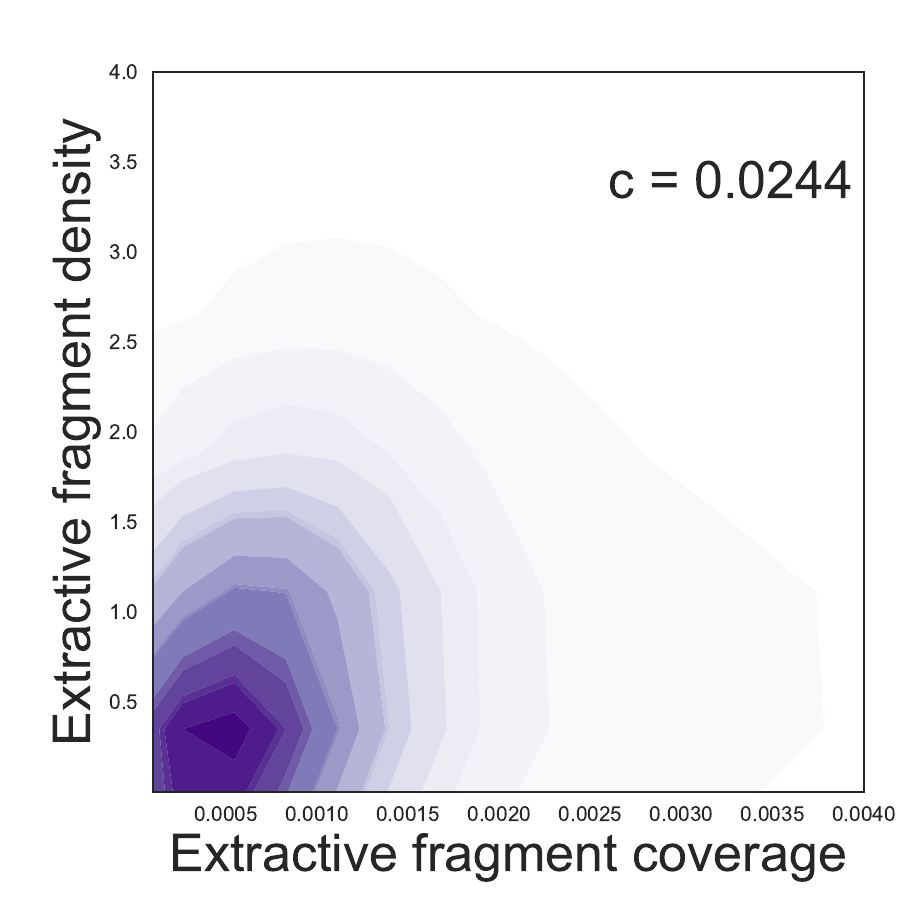}}
          \label{fig_side_GS}
      \end{minipage}%
      \begin{minipage}[t]{0.24\linewidth}
      \centering
          \subfigure[Entities]{
      \includegraphics[width=1.5in]{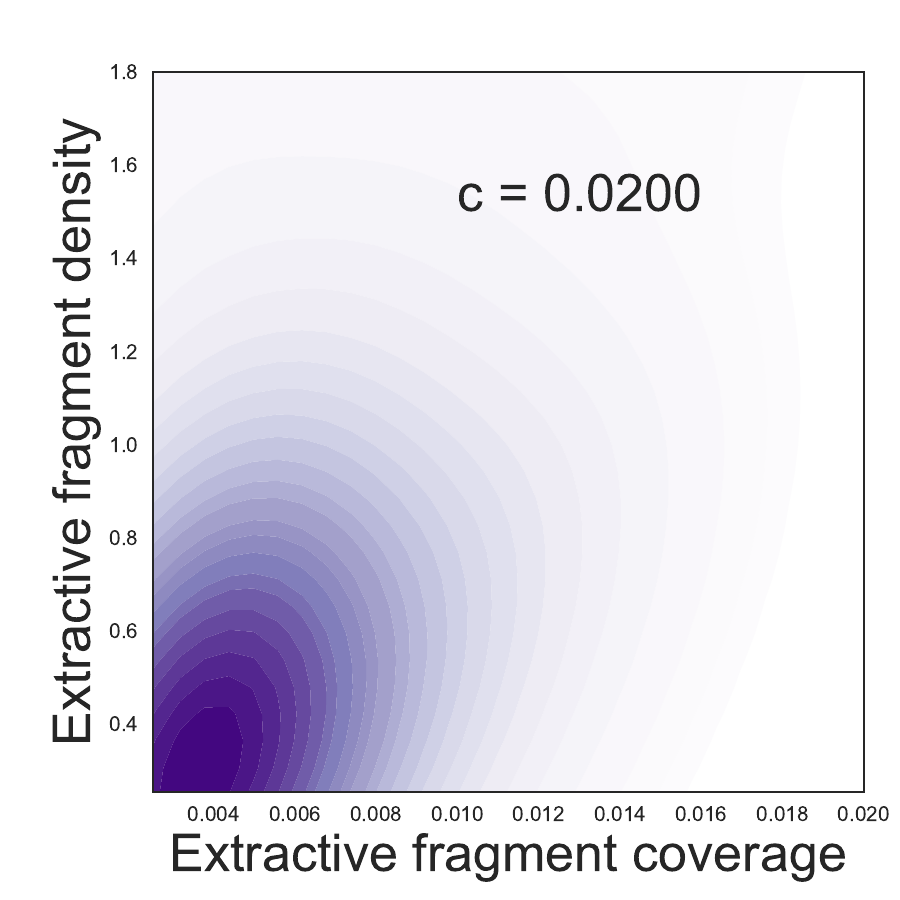}}
          \label{EGAT}
      \end{minipage}
        \begin{minipage}[t]{0.24\linewidth}
        \centering
            \subfigure[CNTLS]{
        \includegraphics[width=1.5in]{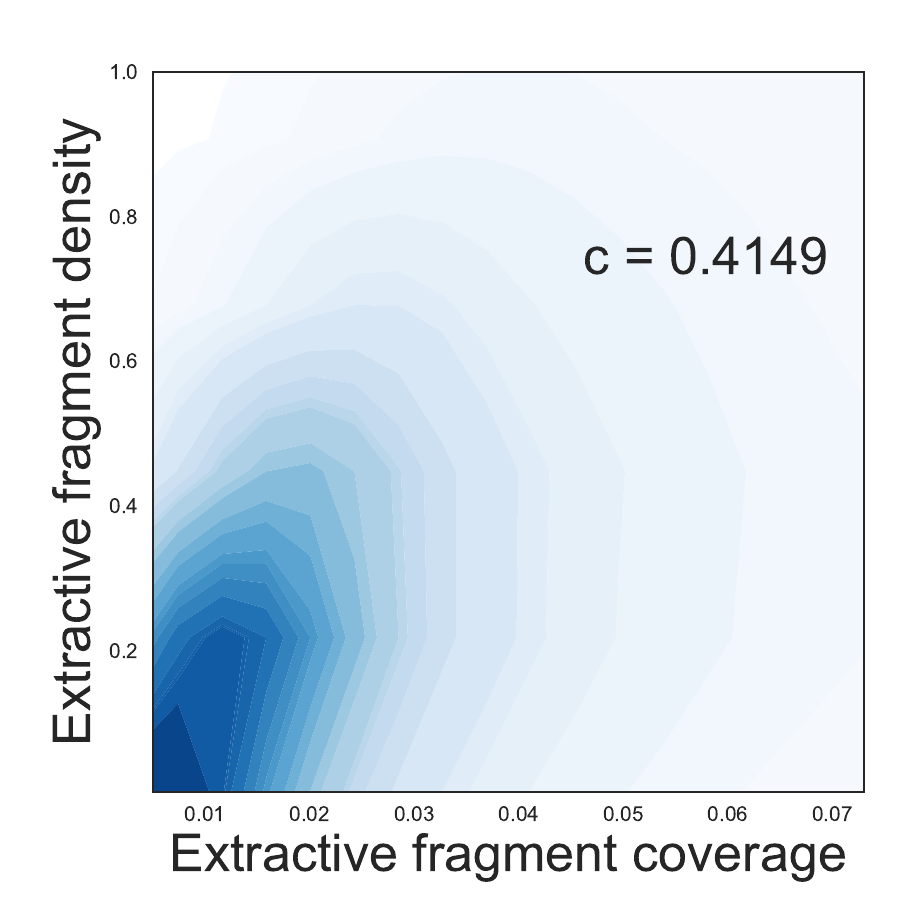}}
            \label{CNTLS2}
        \end{minipage}
      \vspace{-0.15in}
    \caption{Abstractivity\(_{p}\) (p=2) and novel 2-grams distributions on four datasets (subfigures (a), (b), (c), (d)).  Density and coverage distributions of extractive compression scores on four datasets (subfigures (e), (f), (g), (h)). Each box represents a normalized bivariate density plot. 
    The top left corner of each plot shows the median compression ratio \(c\) between summaries and source text.
    }
    \label{distribution_pics}
\end{figure*}

\begin{table*}[htb]
  \footnotesize
  \centering
  \caption{Results of the extractive methods on the constructed CNTLS datasets.}
  \label{extractive_results}
  \renewcommand\arraystretch{1.3}
  \setlength{\tabcolsep}{1.5mm}{
  \begin{tabular}{l|ll|ll|ll|l}
    \toprule 
  \multirow{2}{*}{\textbf{Methods}} & \multicolumn{2}{c|}{\textbf{Concat}}                                 & \multicolumn{2}{c|}{\textbf{Agree}}                           & \multicolumn{2}{c|}{\textbf{Align+ m:1}}                           & \multirow{2}{*}{\textbf{Date F1}}   \\ \cline{2-7}
                           & \multicolumn{1}{c|}{ROUGE-1} & \multicolumn{1}{c|}{ROUGE-2} & \multicolumn{1}{c|}{ROUGE-1} & ROUGE-2               & \multicolumn{1}{c|}{ROUGE-1} & ROUGE-2               &                            \\
                           \midrule
  Oracle Date              & \multicolumn{1}{l|}{.460}        & .190                              &  \multicolumn{1}{l|}{.330}        & \multicolumn{1}{l|}{.159} & \multicolumn{1}{l|}{.039}        & \multicolumn{1}{l|}{.153} &  1.0                          \\ 
  Oracle Text              & \multicolumn{1}{l|}{.454}        & .240                             & \multicolumn{1}{l|}{.309}        & \multicolumn{1}{l|}{.145} & \multicolumn{1}{l|}{.323}        & \multicolumn{1}{l|}{.157} & .993                           \\ 
  Oracle Full              & \multicolumn{1}{l|}{.440}        &   .230                           & \multicolumn{1}{l|}{.320}        & \multicolumn{1}{l|}{.160} & \multicolumn{1}{l|}{.318}        & \multicolumn{1}{l|}{.160} &   1.0                         \\ \midrule
  CLUST                    & \multicolumn{1}{l|}{.224}   & .073                        & \multicolumn{1}{l|}{.031}   & .004                 & \multicolumn{1}{l|}{.035}   & .006                 & \multicolumn{1}{l}{.326} \\ 
  DATAWISE\(^{\texttt{Textrank}}\)                 & \multicolumn{1}{l|}{.359}   & .136                        & \multicolumn{1}{l|}{.154}   & .069                 & \multicolumn{1}{l|}{.174}   & .095                 & \multicolumn{1}{l}{.605} \\ 
  DATAWISE\(^{\texttt{Centro-rank}}\)                  & \multicolumn{1}{l|}{.383}   & .152                        & \multicolumn{1}{l|}{.190}   & .090                 & \multicolumn{1}{l|}{.196}   & .102                 & \multicolumn{1}{l}{.605} \\ 
  DATAWISE\(^{\texttt{Centro-opt}}\)                 & \multicolumn{1}{l|}{.396}   & .157                        & \multicolumn{1}{l|}{.184}   & .089                 & \multicolumn{1}{l|}{.206}   & .104                 & \multicolumn{1}{l}{.605} \\ \bottomrule
  \end{tabular}
  }
  \end{table*}

\subsection{The Abstractivity of CNTLS}
We use density, coverage, and compression to understand the data distribution of constructed corpus.  
Additionally, we showcase the distribution of the samples by combining the values of abstractivity\(_{p}\) (p=2) and novel 2-grams. These graphical representations visually convey the degree of abstractivity in the summaries within the CNTLS corpus:
\begin{itemize}[leftmargin=*]
\item 
The 'Density and coverage distributions' plots exhibit a positive correlation with extractivity, concentrating higher extractivity in the partition positioned around the upper right corner. 
\item The plots illustrating `abstractivity\(_{p}\) and novel 2-grams distributions' demonstrate a positive correlation with abstractivity, with distributions centered closer to the upper right corner, indicating highly abstractive summaries.
\item A higher compression ratio \(c\) increases the summarization task's difficulty, requiring the model to accurately capture crucial aspects or events from the original text for concentrating into a concise and informative summary.
\end{itemize}

Figure~\ref{distribution_pics} illustrates that Crisis and TL17 datasets predominantly prefer abstractive summaries, with the distribution tending to decrease and shift left. However, the distribution covering a wide range of coverage and density shows that CNTLS spans from highly extractive to highly abstractive. 

Based on Tables~\ref{Average_metric_results} and Figures~\ref{distribution_pics}, we can conclude that CNTLS encompasses a broader spectrum of summarization styles, demonstrating significant summary diversity.

\begin{table*}[htb]
  \footnotesize
  \centering
  \caption{Results of the abstractive methods on the constructed CNTLS datasets.}
  \label{abstractive_results}
  \renewcommand\arraystretch{1.3}
  \setlength{\tabcolsep}{1.2mm}{
  \begin{tabular}{l|ll|lc|cc|l}
    \toprule 
  \multirow{2}{*}{\textbf{Methods}} & \multicolumn{2}{c|}{\textbf{Concat}}                                 & \multicolumn{2}{c|}{\textbf{Agree}}                           & \multicolumn{2}{c|}{\textbf{Align+ m:1}}                           & \multirow{2}{*}{\textbf{Date F1}}   \\ \cline{2-7}
                           & \multicolumn{1}{c|}{ROUGE-1} & \multicolumn{1}{c|}{ROUGE-2} & \multicolumn{1}{c|}{ROUGE-1} & ROUGE-2               & \multicolumn{1}{c|}{ROUGE-1} & ROUGE-2               &                            \\
                           \midrule
  Chinese-Alpaca-2-7B-4K              & \multicolumn{1}{l|}{.020}        &   .006                           & \multicolumn{1}{l|}{.015}        & \multicolumn{1}{l|}{.004} & \multicolumn{1}{l|}{.016}        & \multicolumn{1}{l|}{.004} &  .603                          \\ 
  ChatGLM-6B-2K              & \multicolumn{1}{l|}{.262}        &  .052                            & \multicolumn{1}{l|}{.093}        & \multicolumn{1}{l|}{.022} & \multicolumn{1}{l|}{.103}        & \multicolumn{1}{l|}{.024} &  .602                          \\
  ChatGLM2-6B-8K             & \multicolumn{1}{l|}{.248}        &  .052                            & \multicolumn{1}{l|}{.103}        & \multicolumn{1}{l|}{.024} & \multicolumn{1}{l|}{.112}        & \multicolumn{1}{l|}{.025} &    .605  \\
  ChatGLM2-6B-32K  & \multicolumn{1}{l|}{.268}        &  .067                           & \multicolumn{1}{l|}{.111}        & \multicolumn{1}{l|}{.034} & \multicolumn{1}{l|}{.120}        & \multicolumn{1}{l|}{.035} &    .605
   \\ \bottomrule
  \end{tabular}
  }
  \end{table*}

\section{Timeline Summarization Systems}
We evaluate multiple summarization systems to understand the challenges posed by the CNTLS dataset. We assess conventional extractive models and an extractive oracle, providing an upper bound for extractive performance within the corpora. Additionally, we pioneer the use of LLMs in timeline summarization.
For the generative timeline summarization model, we choose LLMs capable of handling lengthy text sequences. While the token sequence length for CNTLS in the 8th column of Table 3 is 554k, most large models have a maximum sequence length of approximately 32K, rendering direct input of the complete token sequence unfeasible.
Therefore, we employ regression methods for date prediction, feeding each time point's respective document into the large model for summary generation.
\subsection{Extractive systems}
\textbf{Event-clust summarization approaches}: Clust~\cite{GhalandariI20} uses DATEMENTIONCOUNT~\cite{GhalandariI20}~\footnote{DATEMENTIONCOUNT: Rank by how often
the cluster date is mentioned throughout the
input collection.} to
rank clusters. It then employs CENTROID-OPT~\cite{DRadevJST04} to rank sentences based on their similarity to the centroid of all sentences for timeline summarization.  

\noindent\textbf{Date-wise summarization approaches}: 
Datewise~\cite{GhalandariI20} employs supervised date selection, PM-MEAN~\cite{GhalandariI20} for candidate selection, and CENTROID-OPT~\cite{DRadevJST04} for timeline summarization. 
\subsection{Abstractive systems}
\textbf{Alpaca for long text summarization}: 
The Chinese-Alpaca-2-7B-4K~\cite{abs-2304-08177} is expanded on the original Alpaca~\cite{Taori2023a} incorporating Chinese vocabulary and undergoing additional pre-training with Chinese data to enhance its foundational semantic understanding in Chinese. 
The model has a context length of 4K with position interpolation, making them suitable for text summarization applications. The implementation of Chinese-Alpaca for summarization inference is done using the Transformer~\footnote{\emph{\url{https://github.com/huggingface/transformers}}}.


\noindent\textbf{ChatGLM for long text summarization}: 
ChatGLM-6B-2K\footnote{\emph{\url{https://github.com/THUDM/ChatGLM-6B}}} and ChatGLM2-6B-2K\footnote{\emph{\url{https://github.com/THUDM/ChatGLM2-6B}}} are an open bilingual language models based on General Language Model (GLM)~\cite{DuQLDQY022} and use LLMs' technology similar to ChatGPT~\cite{Openaichatgpt},. They are  optimized for Chinese Q\&A, dialogue and summarization. 
The implementation of ChatGLM for summarization inference is achieved using the \texttt{HuggingFaceHub}\footnote{\emph{\url{https://huggingface.co/THUDM/chatglm-6b}}}. ChatGLM2-6B-32K further enhances its ability to understand long texts compared to ChatGLM2-6B, allowing it to handle up to a 32K context length.

\section{Timeline Summarization Performances}
\subsection{Evaluation Metrics}
We present ROUGE-1 and ROUGE-2 F1 scores of \textit{concat}, \textit{agreement}, \textit{align+ m:1} metrics, and Date F1-score for timeline summarization systems as discussed in ~\cite{MarkertM17,MartschatM18,GhalandariI20}. These metrics assess concatenation of daily summaries, consideration of matching days, and alignment based on date and content similarity. Date selection is evaluated using the F1 score.

The extractive ORACLE is obtained through a greedy approach. We assess Date-ORACLE, Text-ORACLE, and Full-ORACLE for extractive summarization systems. Date-ORACLE selects the correct (Ground-truth) dates and employs CENTROID-OPT for date summarization. Text-ORACLE uses regression to select dates and constructs a summary for each date by optimizing the ROUGE score with the ground-truth summaries. Full-ORACLE selects the correct dates and generates a summary for each date by optimizing the ROUGE score with the ground-truth summaries.

\subsection{Experimental Results}
According to our experimental results in Table~\ref{extractive_results} and ~\ref{abstractive_results}, while the method ranking remains stable, performance varies significantly across extractive and abstractive methods within datasets. Specifically, the ORACLE predominantly outperforms other systems in extractive methods, indicating that both time prediction and extraction summarization impact extraction method performance. 

In both extractive and generative methods, we use the optimal regression time prediction method to ensure a fair comparison between the extraction and generative methods. As observed in Tables~\ref{extractive_results} and ~\ref{abstractive_results}, the effectiveness of partial extractive methods surpasses that of the generative LLMs (ChatGLM2-7B). Among the generative LLMs used in our experiments, ChatGLM2-6B-32K shows slightly higher scores in ROUGE-1 and ROUGE-2 across all three time prediction methods among all the generative LLMs, indicating that for our timeline summarization task, enlarging the total number of input tokens can be more beneficial (ChatGLM2-6B-2K to ChatGLM2-6B-32K). However, compared to extractive methods, none of the models achieves particularly high ROUGE scores. In other words, the improvements in extractive methods are most evident, far exceeding the performance of large and long contextual generative models.

Moreover, CNTLS, with a high compression rate (41.49, as shown in Table~\ref{Average_metric_results}), generally makes summarization more challenging, and longer topic durations (\texttt{{\small A.Duration}}=4,437, as shown in Table~\ref{FFCC_RANK_results}) impact the performance of generative methods. As previously mentioned, our efforts to create a large model supporting longer contextual inputs result in marginal improvements when increasing the sequence length from 2k to 32k. However, there is no significant difference in summary performance among Chinese-Alpaca and Chinese-ChatGLM2. These results highlight that there is still significant room for improvement in large models for long sequence timeline summary tasks.

\section{Conclusions}
We introduce CNTLS, a dataset comprising articles and their timeline summaries authored by online publication editors. In comparison to existing datasets, ours offers a greater number of topics and a diverse set of summaries. To validate our dataset, we conducted benchmark evaluations using prominent extractive frameworks. Additionally, we pioneer the integration of LLMs in the domain of timeline summarization. Our proposed timeline systems, along with the corresponding analysis of LLMs-based abstractive strategies, open new avenues for assessing the challenges of timeline summarization tasks and for advancing future summarization models. Experimental results in abstractive summarization reveal opportunities for further enhancements in this domain.

\section*{Acknowledgements}
This work is supported in part by the National Natural Science Foundation of China (No.U20B2053).

\bibliography{anthology,custom}
\bibliographystyle{acl_natbib}

\end{document}